\documentclass[10pt,twocolumn,letterpaper]{article}

 \usepackage[pagenumbers]{cvpr} 

\usepackage{mathtools}
\usepackage{subcaption, caption}
\usepackage[sort&compress]{natbib}

\usepackage{booktabs}       
\usepackage{amsfonts}       
\usepackage{nicefrac}       
\usepackage{microtype}      

\usepackage{times}
\usepackage{algorithm,algorithmicx,algpseudocode}
\usepackage{graphicx}
\usepackage{amsmath}
\usepackage{amssymb}

\usepackage{diagbox}
\usepackage{multicol}
\usepackage{enumerate}
\usepackage{times}
\usepackage{epsfig}
\usepackage{threeparttable}
\usepackage{enumitem}
\usepackage{multirow}
\usepackage{color}
\usepackage{array}
\usepackage{setspace}
\usepackage{makecell}

\definecolor{cvprblue}{rgb}{0.21,0.49,0.74}
\usepackage[pagebackref,breaklinks,colorlinks,allcolors=cvprblue]{hyperref}

\def\paperID{} 
\def\confName{CVPR}
\def\confYear{2026}

\title{NEC-Diff: Noise-Robust Event–RAW Complementary Diffusion for Seeing Motion in Extreme Darkness}
\author{Haoyue Liu\textsuperscript{1}\footnotemark[2], Jinghan Xu\textsuperscript{1}\footnotemark[2], Luxin Feng\textsuperscript{1}, Hanyu Zhou\textsuperscript{2}, Haozhi Zhao\textsuperscript{1}, Yi Chang\textsuperscript{1}\footnotemark[1], Luxin Yan\textsuperscript{1}\\
	\textsuperscript{1} National Key Lab of Multispectral Information Intelligent Processing Technology\\
	School of Artificial Intelligence and Automation, Huazhong University of Science and Technology\\
	\textsuperscript{2} School of Computing, National University of Singapore \\
	{\tt\small \{liuhy, xujinghan, fengluxin, yichang\}@hust.edu.cn, hy.zhou@nus.edu.sg}}

\begin{document}
\maketitle
{
	\renewcommand{\thefootnote}%
	{\fnsymbol{footnote}}
	\footnotetext[0]{$\dagger$Equal contribution. *Corresponding author.} 
}
\begin{abstract}
High-quality imaging of dynamic scenes in extremely low-light conditions is highly challenging. Photon scarcity induces severe noise and texture loss, causing significant image degradation. Event cameras, featuring a high dynamic range (120 dB) and high sensitivity to motion, serve as powerful complements to conventional cameras by offering crucial cues for preserving subtle textures. However, most existing approaches emphasize texture recovery from events, while paying little attention to image noise or the intrinsic noise of events themselves, which ultimately hinders accurate pixel reconstruction under photon-starved conditions. In this work, we propose \textbf{NEC-Diff}, a novel diffusion-based event–RAW hybrid imaging framework that extracts reliable information from heavily noisy signals to reconstruct fine scene structures. The framework is driven by two key insights: (1) combining the linear light-response property of RAW images with the brightness-change nature of events to establish a physics-driven constraint for robust dual-modal denoising; and (2) dynamically estimating the SNR of both modalities based on denoising results to guide adaptive feature fusion, thereby injecting reliable cues into the diffusion process for high-fidelity visual reconstruction. Furthermore, we construct the \textbf{REAL} (Raw and Event Acquired in Low-light) dataset which provides 47,800 pixel-aligned low-light RAW images, events, and high-quality references under 0.001–0.8 lux illumination. Extensive experiments demonstrate the superiority of NEC-Diff under extreme darkness. The project are available at: \url{https://github.com/jinghan-xu/NEC-Diff}.

\end{abstract}
    
\section{Introduction}
\label{sec:intro}
In extremely dark environments, the limited number of photons reaching the sensor causes conventional cameras to suffer from severe noise and loss of texture details. Although extending the exposure time can collect more photons to improve image quality, it inevitably leads to motion blur in dynamic scenes. Increasing camera gain amplifies both signal and noise, further degrading the signal-to-noise ratio (SNR). Consequently, achieving high-quality imaging of dynamic scenes under extremely low-light conditions remains highly challenging.

\begin{figure}[t]
	\centering
	\includegraphics[width=1.0\linewidth]{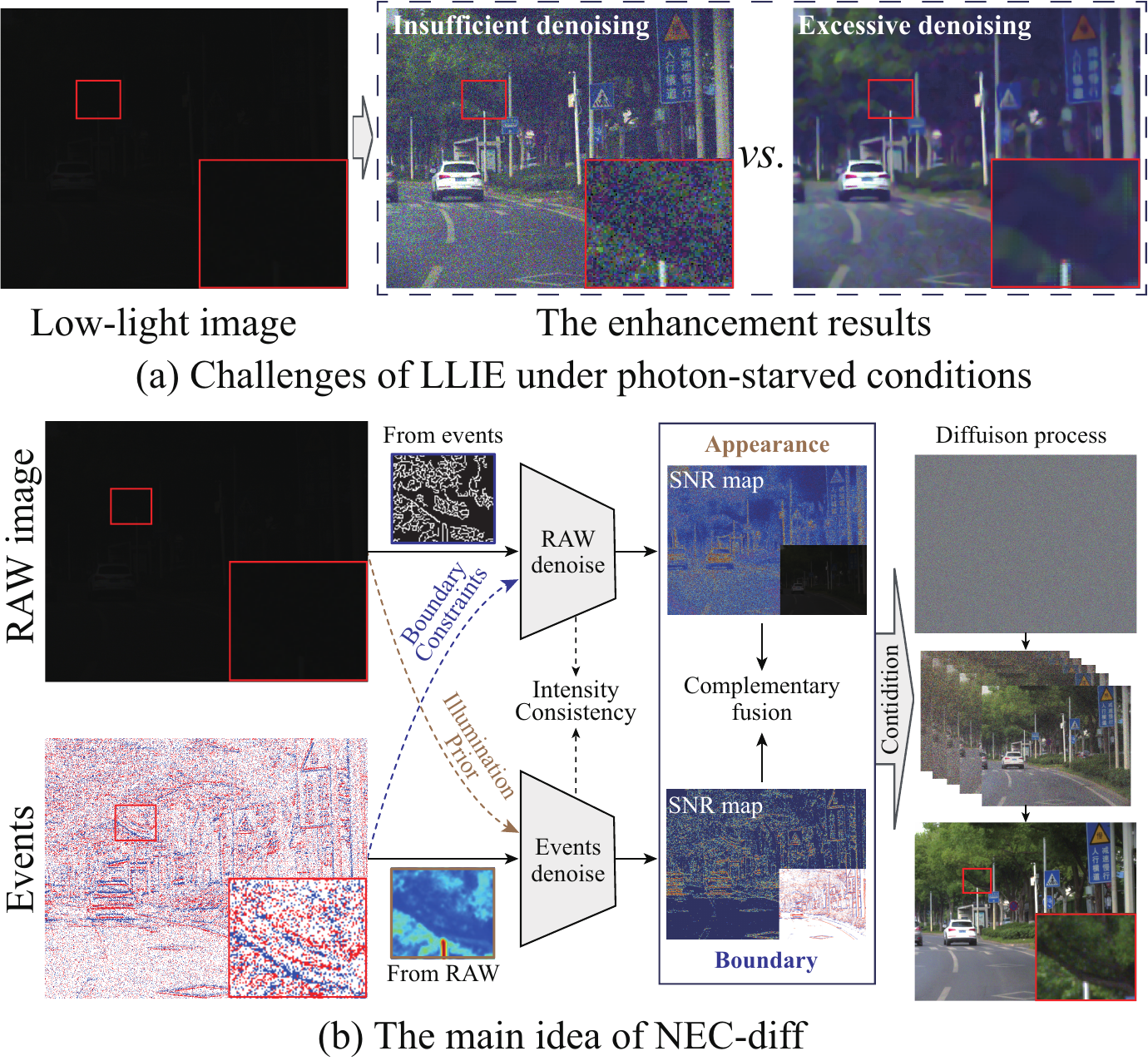}
	\caption{Illustration of problem and main idea. (a) LLIE methods suffer from a trade-off between \textbf{texture preservation} and \textbf{noise suppression}. (b) Events effectively complement textures but introduce additional noise. NEC-Diff exploits the characteristics of both events and RAW images to achieve robust denoising while preserving textures, fusing features guided by SNR and injecting them into the diffusion model to achieve high-fidelity results.}
	\label{fig1_main_idea}
\end{figure}

Existing low-light image enhancement (LLIE) methods \cite{ma2022toward,wu2022uretinex,xu2022snr,cao2023physics,liang2023iterative,cai2023retinexformer,wang2024zero} have made notable progress in general low-light scenarios, but they still face significant limitations under extreme darkness and exposure constraints. On one hand, textures that have been lost cannot be faithfully recovered; on the other, achieving a balance between noise suppression and texture preservation is difficult, often resulting in residual noise or oversmoothed details, as shown in Fig~\ref{fig1_main_idea} (a).

To improve noise modeling, several studies utilize RAW images \cite{chen2019seeing,jiang2019learning,gu2019self,wei2020physics,zhu2020eemefn,jin2023dnf,jiang2025learning} as input. Compared with sRGB images, RAW data preserves more unprocessed information and avoids the nonlinear transformations of the image signal processing (ISP) pipeline, thereby facilitating more accurate noise modeling and improved enhancement quality. However, RAW imaging still follows the global exposure paradigm and thus cannot fundamentally solve the information loss caused by short exposure in dynamic scenes, limiting its performance.

Event cameras \cite{lichtsteiner2008128,gallego2020event}, which asynchronously respond to illumination changes, feature a high dynamic range (120 dB) and microsecond-level temporal resolution. They are highly sensitive to motion edges and can effectively complement texture information missing in conventional frames under fast motion. Consequently, many works \cite{jiang2023event,liu2023low,she2025exploring,guo2025eretinex,chenevent} fuse sRGB images and events to improve low-light imaging. Some methods further mitigate image noise via motion consistency \cite{liang2023coherent} and SNR estimation \cite{liang2024towards,chen2025evlight}, or suppress both event and image noise through filtering \cite{kim2024towards,sun2025low}. However, such filtering or single-network solutions struggle to achieve precise denoising while preserving weak signals, thereby limiting imaging quality under extreme darkness. EvRAW \cite{zheng2025evraw} introduces an event–RAW hybrid approach focusing on detail and color recovery, yet it pays limited attention to sensor noise. The severe noise present in both event and frame modalities under low-light conditions raises a fundamental question: \textit{how can we effectively remove noise from dual-modal degraded signals to restore fine scene details? }

To address these issues, we propose a diffusion-based event–RAW hybrid imaging framework, NEC-Diff. Our design focuses on two key aspects: (1) modeling and suppressing noise in severely degraded signals, and (2) efficiently fusing dual-modal features for high-fidelity imaging. Unlike previous methods \cite{jiang2023event,liu2023low,she2025exploring,guo2025eretinex,chenevent,liang2023coherent,liang2024towards,chen2025evlight,kim2024towards,sun2025low}, we adopt RAW images as input, which provide richer scene information and more tractable noise distributions. The NEC-Diff framework consists of three modules: Event–RAW Collaborative Noise Suppression (ECNS), SNR-Guided Reliable Information Extraction (SRIE), and Cross-Modal Attentive Diffusion (CAD).

The main idea of NEC-diff is illustrated in Fig.~\ref{fig1_main_idea} (b). First, to tackle the severe noise in both modalities, the key insight of the ECNS is to exploit the complementary strengths of events and frames to mutually assist in denoising. Since RAW images are linearly correlated with illumination, while event data are dominated by photon shot noise under low-light conditions \cite{guo2022low,graca2021unraveling}, the illumination prior provided by RAW images can effectively guide event denoising. Meanwhile, the bottleneck in image denoising arises from the challenge of distinguishing signal from noise in weakly textured regions; denoised events, in turn, provide high-dynamic-range edge cues, aiding noise suppression in images without oversmoothing subtle textures. Furthermore, we design an intensity consistency loss based on the physical relationship between clean RAW images and corresponding events to better constrain the denoising process. Events exhibit high dynamic range and strong responses to motion edges and local variations, whereas images provide stable global brightness and texture information. To fully exploit their complementary advantages, we design the SRIE to robustly extract reliable information from both modalities by dynamically selecting features from the modality with higher SNR. The CAD further leverages cross-modal attention to fuse these reliable features and injects them as conditional inputs into a diffusion model, enabling robust and high-quality reconstruction in extremely low-light conditions.

Since existing datasets \cite{jiang2023event,liang2024towards,chen2025evlight,kim2024towards,sun2025low,duan2025eventaid} provide only sRGB data, and RealRE \cite{zheng2025evraw} offers event–RAW pairs but lacks low-light cases, we build a coaxial imaging system and construct the Raw and Event Acquired in Low-light (REAL) dataset. REAL contains pixel-aligned triplets of low-light RAW images, low-light events, and high-quality sRGB GTs, with illumination levels ranging from 0.01 lux to 0.8 lux. Overall, Our main contributions are summarized as follows:

\begin{itemize}[leftmargin=10pt]

	\item We provide a novel solution to the noise–texture trade-off inherent in photon-starved imaging, leveraging the texture cues from events and the illumination-consistency relationship between RAW images and events to effectively disentangle sensor noise and preserve fragile signals in extremely dark environments.

	\setlength{\itemsep}{5pt}
	\item By examining the complementary signal behaviors of events and RAW images—events deliver high-fidelity edge responses and RAW images preserve holistic brightness, we introduce NEC-Diff, which adaptively selects reliable cross-modal features based on their SNR and injects them into a diffusion model, achieving high-quality imaging under extreme darkness.
	
	\setlength{\itemsep}{5pt}
	\item We construct the REAL dataset comprising pixel-aligned RAW, event, and sRGB triplets, serving as a valuable benchmark for advancing low-light imaging research. Extensive experiments demonstrate that NEC-Diff achieves state-of-the-art performance across multiple datasets.
	
\end{itemize}

\begin{figure*}
	\centering
	\includegraphics[width=1.0\linewidth]{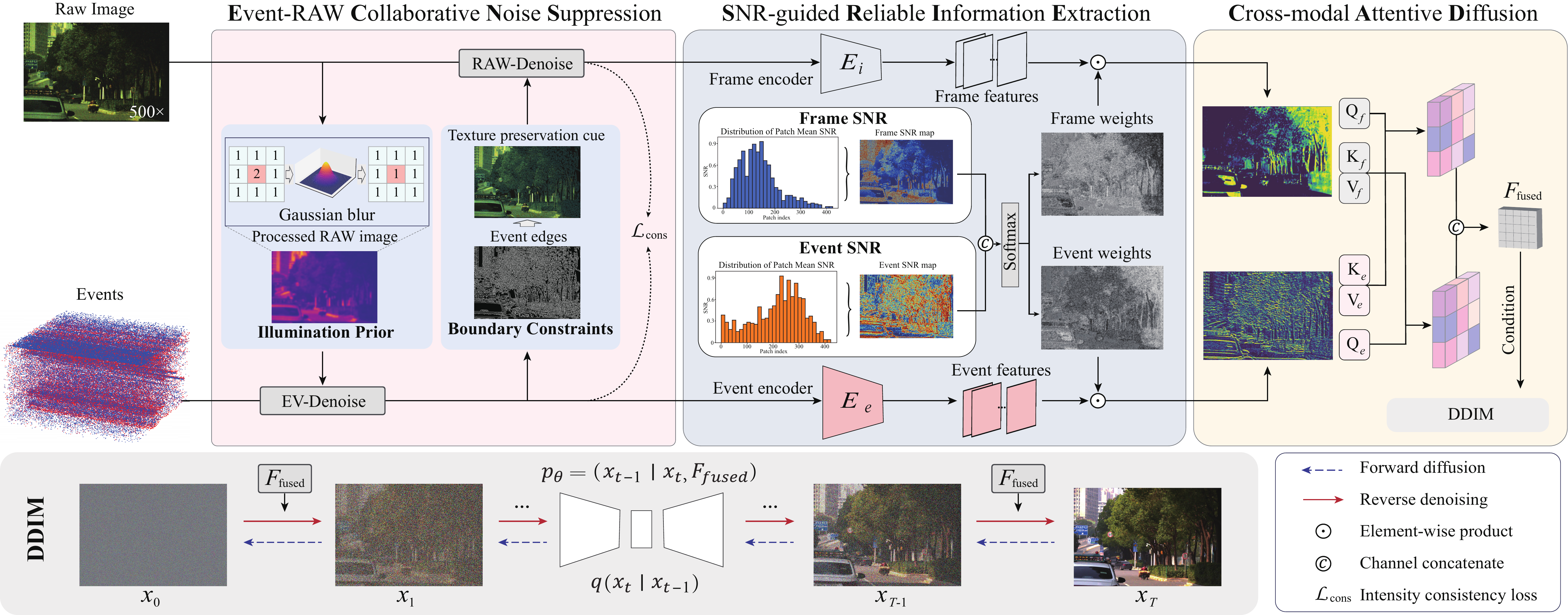}
	\caption{The architecture of NEC-diff includes an Event-RAW Collaborative Noise Suppression (ECNS), a SNR-Guided Reliable Information Extraction (SRIE), and a Cross-Modal Attentive Diffusion (CAD). The ECNS jointly exploits illumination priors from RAW images and texture cues from events for cross-modal denoising. The SRIE adaptively selects high-SNR features from both modalities. The CAD integrates reliable features via cross-modal attention into a diffusion model for high-quality reconstruction.}
	\label{fig2_framework}
\end{figure*}

\section{Related Work}
\label{sec:related_work}

\noindent
\textbf{Frame-based Low-light Imaging.}
Frame-based low-light imaging techniques, such as low-light image enhancement (LLIE), have been extensively explored. According to the input type, LLIE methods can be broadly categorized into two types: sRGB-based \cite{ma2022toward,wu2022uretinex,xu2022snr,cao2023physics,liang2023iterative,cai2023retinexformer,wang2024zero} and RAW-based \cite{chen2019seeing,jiang2019learning,gu2019self,wei2020physics,zhu2020eemefn,jin2023dnf,jiang2025learning} approaches. Compared with RAW images, sRGB images undergo a series of visual corrections in the ISP pipeline, making them more consistent with human visual perception. However, under extremely low-light conditions, the SNR of images drops sharply, and the ISP pipeline instead becomes a burden—both the original signals and noise are distorted by nonlinear transformations, making it more difficult to separate signal from noise during enhancement. Hence, RAW images serve as a more suitable input for extreme low-light enhancement. Single-stage methods \cite{chen2018learning,gu2019self,lamba2021restoring} directly learn a mapping from noisy RAW data to clean sRGB images, while multi-stage methods \cite{zhu2020eemefn,dong2022abandoning,huang2022towards,jin2023dnf} decouple this mapping into multiple nonlinear transformation processes, further improving the final imaging quality.

Although existing RAW-based methods have achieved promising results, the short exposure required for imaging dynamic scenes further reduces the number of photons captured by the sensor, making it difficult to recover reasonable results when the signal is severely lacking.

\noindent
\textbf{Event-based Low-light Imaging.}
Benefiting from the high dynamic range and temporal resolution of event cameras, directly reconstructing images from events has become an efficient approach \cite{rebecq2019events,wang2019event,zhang2020learning,liu2023sensing}. DVS-Dark \cite{zhang2020learning} attempts to improve nighttime performance through domain adaptation, transferring knowledge learned from daytime data. NER-Net \cite{liu2024seeing,liu2025ner} explicitly models the non-ideal event responses in darkness—such as trailing effects and noise—and adopts a low threshold setting to further enhance low-light reconstruction. Nevertheless, since events only record changes in scene brightness, they cannot accurately recover intensity information in smooth regions where no events are triggered.

Recently, event–frame hybrid methods demonstrates strong potential for high-precision imaging under low-light conditions. Several methods \cite{jiang2023event,she2025exploring,guo2025eretinex} directly fuse features from events and frames; however, these approaches often overlook the impact of sensor noise on low-light image quality, leading to unstable fusion results in ultra-dark scenes. EvLowLight \cite{liang2023coherent} reduces the interference of image noise from a motion consistency perspective, while EvLight \cite{liang2024towards} utilizes events to supplement regions with low signal-to-noise ratio in images, yet neither addresses event noise. ELEDNet \cite{kim2024towards} and RETINEV \cite{sun2025low} further employ low-pass filtering and CNNs to process event noise. However, simple low-pass filtering or standalone event-denoising networks struggle to simultaneously preserve weak signals and achieve accurate noise suppression, thereby limiting their generalization and robustness in extremely low-light environments.

This work integrates the illumination priors of RAW images with the texture cues of events for collaborative denoising, where SNR-guided fusion and diffusion-based reconstruction enable robust imaging under extreme darkness.

\section{Event–RAW Complementary Diffusion}
\label{sec3:method}
\subsection{Framework Overview}
\label{sec3.1:overview}

The core challenge of imaging in photon-starved environments lies in robustly suppressing noise while preserving the weak textures of the scene. We propose the Event–RAW Complementary Diffusion framework (Fig.~\ref{fig2_framework}), integrating noise suppression, complementary fusion, and diffusion-based enhancement. We first analyze noise properties under photon-starved conditions, using RAW illumination priors to guide event denoising. Denoised high-dynamic event edges then facilitate texture–noise separation in frames, mitigating over-smoothing. We further use dual-modality SNR information to guide robust cross-modal fusion. Finally, a diffusion model reconstructs high-quality outputs, providing strong robustness and detail preservation in extreme darkness.

\subsection{Event-RAW Collaborative Noise Suppression}
\label{sec3.2:denoise}
\begin{figure}[t]
	\centering
	\includegraphics[width=1.0\linewidth]{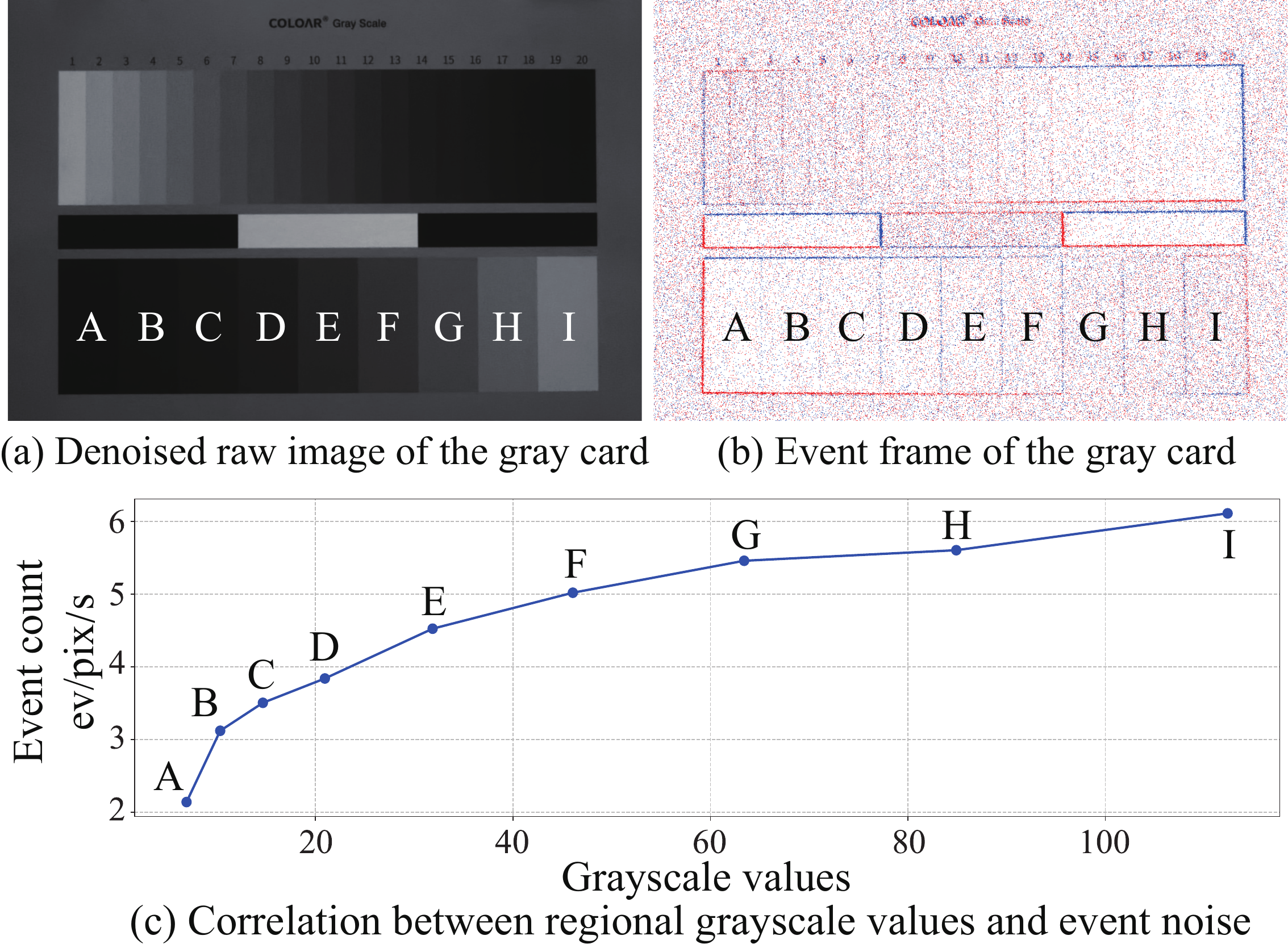}
	\caption{Correlation between event noise density and illumination under low-light conditions. (a) Denoised RAW image indicating illumination intensity across different regions. (b) Event noise density under varying illumination levels. (c) Statistical analysis of event noise density across regions of the gray card.}
	\label{fig3_motivation_noise}
\end{figure}

\textbf{Illumination-guided Event Denoising.} Under low-light conditions, shot noise becomes the dominant source of background activity (BA) in event cameras, with a density more than 50 times higher than other types of noise \cite{guo2022low,graca2021unraveling}. Such noise events approximately follow a Poisson distribution and are closely correlated with illumination, as illustrated in Fig.~\ref{fig3_motivation_noise}. Under a 0.5 lux lighting condition, we simultaneously captured the same grayscale chart using both a frame camera and an event camera. It can be observed that the density of event noise increases with illumination intensity, showing a positive correlation between the two. Meanwhile, RAW images without ISP processing exhibit a linear response with ambient illumination. This observation inspires us to leverage the illumination prior provided by RAW images to guide the event denoising process. To this end, we design an event denoising network similar to EDformer \cite{jiang2024edformer}, which takes the RAW image and raw events as inputs and outputs denoised events. Considering that low-light RAW images inherently contain noise, we employ a simple and efficient Gaussian blur to suppress it. We avoid using more sophisticated denoising methods since the event denoising process only requires a coarse indication of regional illumination rather than precise denoising results.

\noindent
\textbf{Event-assisted Image Denoising.} Image denoising under low-light conditions is challenging because it requires preserving weak signals while removing noise. Event cameras can capture brightness changes with high temporal precision and are highly sensitive to motion edges and structural details, providing reliable edge priors for image denoising. We leverage the high-dynamic edge information from denoised events to guide the image denoising process, suppressing noise while maintaining texture and structural integrity, thus achieving a more balanced denoising performance under extremely low-light conditions. We devise an event-assisted module building upon the architecture of \cite{li2025noise}, 

\noindent
\textbf{Intensity Consistency.} To ensure the effectiveness of both event and frame denoising modules, we further introduce an intensity consistency constraint between RAW images and events, designed through physical modeling of the signal formation process. We first model their respective noise generation processes. The RAW image captured by the frame camera can be expressed as \cite{wei2020physics}:
\begin{equation}
	\setlength\abovedisplayskip{3pt}
	\setlength\belowdisplayskip{3pt}
	R = K I + K N_p + N_{\text{read}} + N_d,
\end{equation}
where, $R$ denotes the pixel value of the RAW image, $K$ is the overall gain, and $I$ represents the number of photoelectrons proportional to the illumination. $N_p$, $N_{\text{read}}$, and $N_d$ denote the photon shot noise, readout noise, and quantization noise, respectively. Based on this model, we denote the noise-free signal component of the RAW image by $\tilde{R} = K I$.

Under extremely low-light conditions, the observed raw discrete event stream $E(t)$ is modeled as~\cite{cao2025noise2image}:
\begin{equation}
	\setlength\abovedisplayskip{3pt}
	\setlength\belowdisplayskip{3pt}
	E(t) = \frac{1}{C} \log \frac{I(t) + b_{pr}}{I(t - \Delta t) + b_{pr}},
\end{equation}
where $b_{pr}$ is the photoreceptor bias term, and $C$ is the event contrast threshold.
However, the ideal condition for an event camera to generate an event at pixel $  (x, y)  $ is given by \cite{paredes2021back}:
\begin{equation}
	\setlength\abovedisplayskip{3pt}
	\setlength\belowdisplayskip{3pt}
	\log I(x, y, t) - \log I(x, y, t - \Delta t) = p C,
\end{equation}
where $p \in \{-1, 1\}$ is the event polarity. For brevity, the spatial coordinates $(x,y)$ are omitted hereafter.
From this, the ideal accumulated event stream $\tilde{E}(t)$ is derived as:
\begin{equation}
	\setlength\abovedisplayskip{3pt}
	\setlength\belowdisplayskip{3pt}
	\tilde{E}(t) = \frac{1}{C} \log \frac{K I(t)}{K I(t - \Delta t)} = \frac{1}{C} \log \frac{\tilde{R}(t)}{\tilde{R}(t - \Delta t)}.
	\label{eq:consistency}
\end{equation}

Building on the physical models above, we jointly optimize the denoising modules by enforcing strict intensity consistency between their outputs. Specifically, let $\hat{R}(t)$ and $\hat{E}(t)$ denote the denoised RAW image and refined event stream, respectively. Ideally, these predictions should satisfy the logarithmic relationship derived in Eq.~\ref{eq:consistency}. To leverage this intrinsic physical correspondence as a complementary constraint, we introduce an intensity consistency loss computed directly from the network outputs:
\begin{equation}
	\setlength\abovedisplayskip{3pt}
	\setlength\belowdisplayskip{3pt}
	\mathcal{L}_{\text{cons}} = \left\| \hat{E}(t) \cdot C - \log \frac{\hat{R}(t) + \epsilon}{\hat{R}(t - \Delta t) + \epsilon} \right\|_1,
\end{equation}
where $C$ represents a learnable scaling factor adapting the physical contrast threshold, and $\epsilon$ is a small constant to prevent numerical instability. This loss enforces that the brightness change between consecutive denoised frames matches the event amplitudes, ensuring physical consistency and structural alignment across the two modalities.

\begin{figure}
	\centering
	\includegraphics[width=1.0\linewidth]{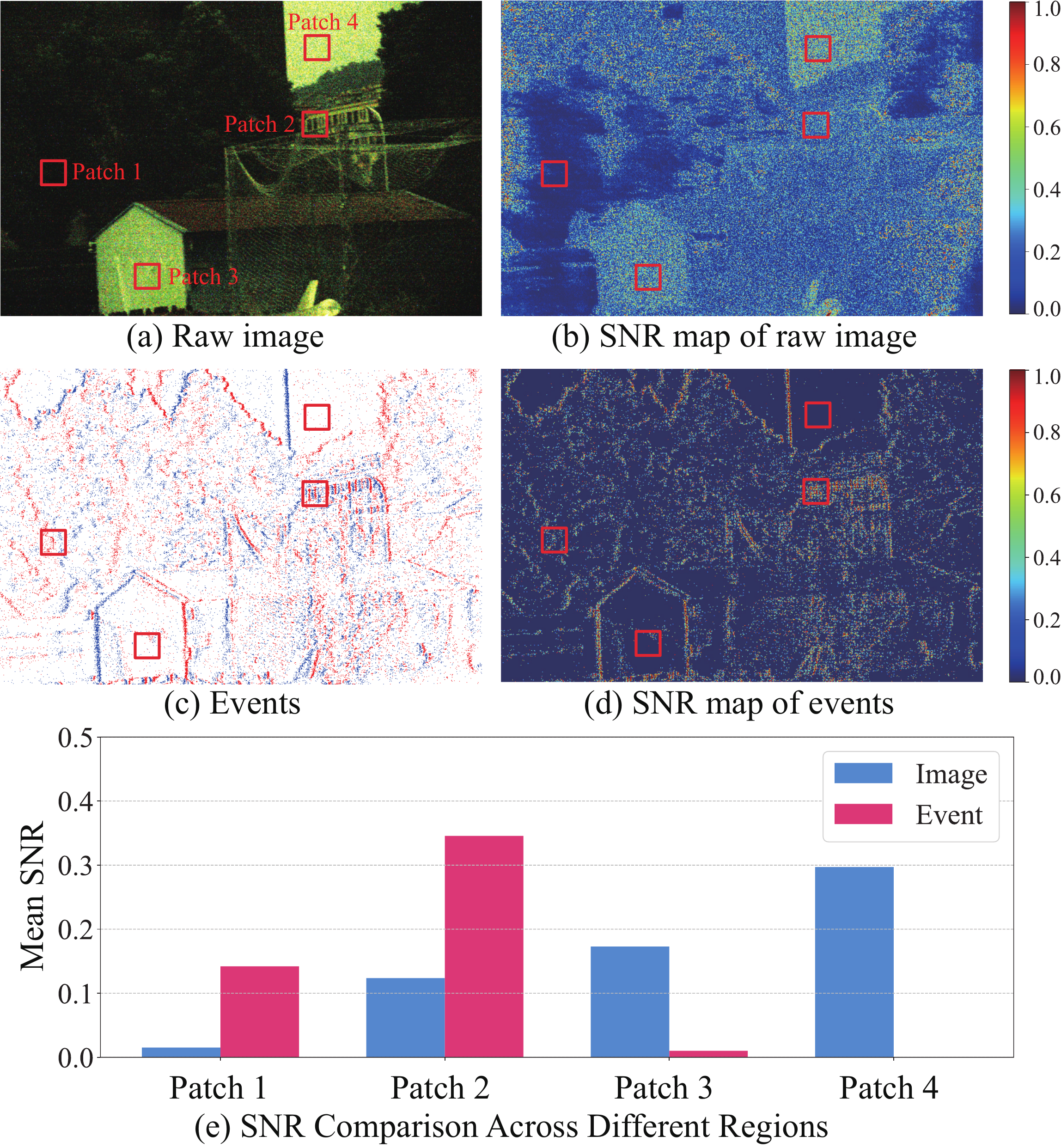}
	\caption{SNR Comparison between RAW and Event Modalities. (a) and (c) show the visualizations of the RAW image and events, while (b) and (d) present their SNR maps. (e) compares the SNR of the image and events across different regions. It can be observed that in dark regions with rich textures, the event SNR is higher, whereas in smooth areas, the event SNR approaches zero.}
	\label{fig4_motivation_snr}
\end{figure}

\begin{figure*}
	\centering
	\includegraphics[width=1.0\linewidth]{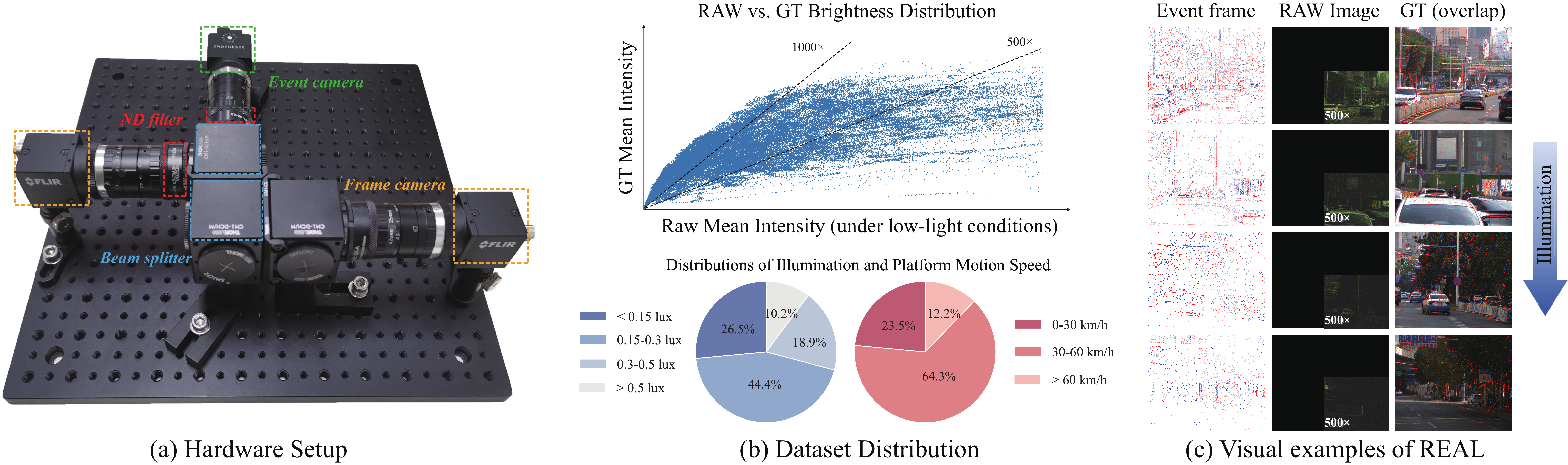}
	\caption{Details of the REAL dataset. (a) Hardware setup of the coaxial imaging system. (b) Distributions of image-pair brightness and platform motion speed. (c) Visualization of representative data pairs.}
	\label{fig5_dataset}
\end{figure*}

\noindent
\subsection{SNR-Guided Reliable Information Extraction}
\label{sec3.3:fusion}
Under extremely low-light conditions, the image and event modalities exhibit notable differences in signal reliability across spatial regions. As illustrated in Fig.~\ref{fig4_motivation_snr}, the SNR of the image is mainly affected by illumination, with brighter regions exhibiting higher signal quality. In contrast, events are more reliable around edge and texture regions but tend to be absent in smooth areas. Compared with popular hybrid LLIE methods such as EvLight \cite{liang2024towards}, which solely relies on image SNR as a weighting basis to supplement information in low-SNR image regions with events, we simultaneously consider the SNR of both modalities during information extraction, effectively mitigating information loss. For instance, in dark smooth-textured regions where events are nearly absent, greater emphasis should be placed on the weak signals provided by the image.

To implement this, we propose a SNR-guided reliable information extraction module, which models the signal reliability of both image and event modalities. Specifically, we use the denoised outputs from the previous stage to compute SNR maps for both image and event modalities. Let $M_{\text{in}}$ denote the original input and $M_{\text{den}}$ the denoised result; then the SNR map for each modality is defined as:
\begin{equation}
	\setlength\abovedisplayskip{3pt}
	\setlength\belowdisplayskip{3pt}
	M_{\text{SNR}} = 10 \cdot \log \frac{M_{\text{in}}^2}{(M_{\text{in}} - M_{\text{den}})^2 + \epsilon},
\end{equation}

The resulting SNR maps provide an intuitive measure of signal reliability at each spatial location. 
We then process the image and event SNR maps jointly using a lightweight network, 
and apply a channel-wise softmax to generate the final spatial weight maps $W_{\text{img}}$ and $W_{\text{evt}}$. 
The normalized weight maps are then applied to the corresponding modality features via element-wise multiplication:
\begin{equation}
	\setlength\abovedisplayskip{3pt}
	\setlength\belowdisplayskip{3pt}
	F_{\text{img-w}} = F_{\text{img}} \odot W_{\text{img}}, \quad 
	F_{\text{evt-w}} = F_{\text{evt}} \odot W_{\text{evt}}.
\end{equation}

This mechanism allows the network to focus on the high-confidence regions of each modality while adaptively suppressing noise-dominated areas, thereby improving robustness and detail fidelity in low-light scenarios. Building on this, we further construct a cross-modal fusion mechanism in the next section to fully exploit the complementary information from both event and image modalities.

\subsection{Cross-Modal Attentive Diffusion}
\label{sec3.4:DDIM}
In Sec.~\ref{sec3.3:fusion}, we extract reliable features from both image and event modalities using the SNR-guided mechanism. Event features respond strongly to dynamic edges and local detail variations, while image features provide more stable global brightness and texture information. To fully leverage the complementary strengths of these two modalities, we design a Cross-modal attentive diffusion module to achieve adaptive feature fusion and reconstruction.

Specifically, the weighted image features $F_{\text{img-w}}$ and weighted event features $F_{\text{evt-w}}$ obtained in the previous stage are fed into a bidirectional cross-modal attention module for feature interaction:
\begin{align}
	\setlength\abovedisplayskip{3pt}
	\setlength\belowdisplayskip{3pt}
	\resizebox{0.9\hsize}{!}{$A_E = \textit{Softmax}\Big(\frac{Q_I K_E^\top}{\sqrt{d}}\Big) V_E, A_I = \textit{Softmax}\Big(\frac{Q_E K_I^\top}{\sqrt{d}}\Big) V_I$},
\end{align}
where $Q$, $K$, and $V$ denote queries, keys, and values, and $d$ is the feature dimension. 
Through this bidirectional interaction, the event modality supplements image features with details and dynamic responses at motion edges, 
while the image modality provides stable absolute brightness distributions to guide event features in flat regions. This ensures a balanced representation that preserves both global brightness consistency and local detail fidelity.

The interacted features are then concatenated along the channel dimension to obtain a unified multi-modal representation $F_{\text{fused}}$, 
which preserves structural information and illumination priors while forming a joint feature suitable for the diffusion model.

During the diffusion-based reconstruction stage, the fused multi-modal feature $F_{\text{fused}}$ serves as a conditional input to guide the noise prediction network $\epsilon_\theta$, 
which predicts the noise at each timestep based on the current state $x_t$:
\begin{equation}
	\setlength\abovedisplayskip{3pt}
	\setlength\belowdisplayskip{3pt}
	\hat{\epsilon}_\theta = \epsilon_\theta(x_t, F_{\text{fused}}, t).
\end{equation}

The image is then progressively reconstructed following the deterministic DDIM sampling process:
\begin{equation}
	\setlength\abovedisplayskip{3pt}
	\setlength\belowdisplayskip{3pt}
	x_{t-1} = \text{DDIM}(x_t, \hat{\epsilon}_\theta, \alpha_t),
\end{equation}
By employing this deterministic sampling strategy, the model substantially reduces diffusion steps while maintaining stable reconstruction. The fused cross-modal features offer strong conditioning cues, enabling the diffusion model to progressively recover the true image distribution in low-SNR regions and achieve a balanced improvement in physical consistency, structural integrity, and texture fidelity.

\begin{figure*}
	\centering
	\includegraphics[width=1.0\linewidth]{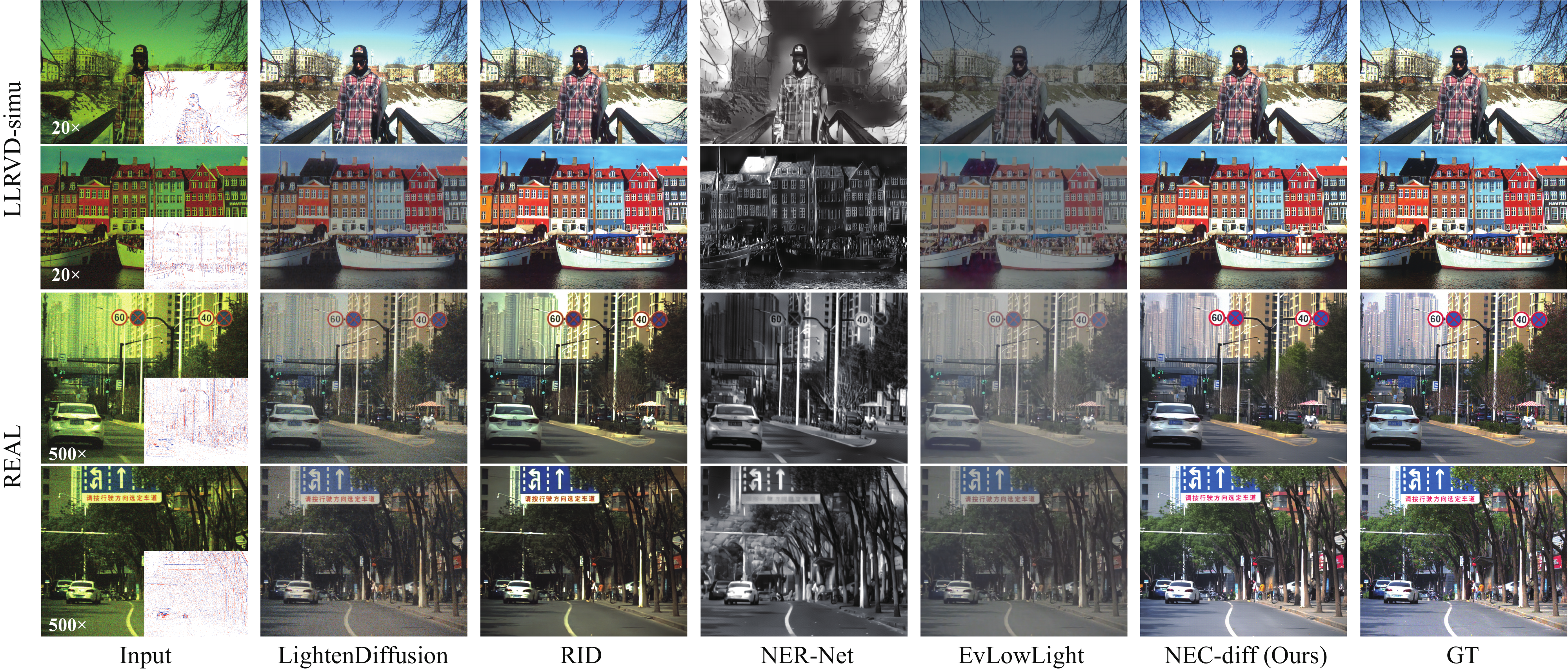}
	\caption{Visual comparison between other SOTA methods and the proposed NEC-diff across different datasets.}
	\label{fig6_qualitive}
\end{figure*}

\subsection{Optimization}
\label{sec3.5:Optimization}
\noindent
\textbf{Loss Function.} We adopt a two-stage training strategy. In the first stage, the image and event noise suppression modules are trained independently, without introducing the other modality, to reduce the training burden for the subsequent DDIM stage. The dataset and training configurations follow \cite{jiang2024edformer,li2025noise} . In the second stage, cross-modal consistency constraints are incorporated for joint training. Specifically, three loss terms are employed: the pixel reconstruction loss $\mathcal{L}_{\text{rec}}$, the gradient-preserving loss $\mathcal{L}_{\text{grad}}$, and the physical consistency loss $\mathcal{L}_{\text{cons}}$ defined previously. The overall loss is formulated as:
\begin{equation}
	\setlength\abovedisplayskip{3pt}
	\setlength\belowdisplayskip{3pt}
	\mathcal{L}_{\text{total}} = \mathcal{L}_{\text{rec}} + \lambda_{\text{grad}} \mathcal{L}_{\text{grad}} + \lambda_{\text{cons}} \mathcal{L}_{\text{cons}},
\end{equation}
where $\lambda_{\text{grad}}=10$ and $\lambda_{\text{cons}}=0.5$ are hyperparameters balancing the contributions of each term. The consistency loss $\mathcal{L}_{\text{cons}}$ ensures that the ECNS module produces reliable pre-denoised outputs, while the reconstruction and gradient losses guide the network to match ground truth in both intensity and edge structures.

\noindent
\textbf{Implementation Details.} We optimize all model parameters using the Adam optimizer with an initial learning rate of $1\times10^{-4}$ for 50 epochs. Input images are cropped to $256 \times 256$ patches before being fed into the network. The forward diffusion process is set to 1000 steps. All experiments are conducted on a single NVIDIA RTX 4090 GPU.

\section{REAL Dataset}
Capturing aligned visual data under extremely low-light and dynamic scenes is challenging, especially when pursuing precise spatial correspondence, reliable timing, and physically accurate dark-scene responses. To address this, we design a coaxial multi-sensor imaging system and emulate ultra-low illumination in real outdoor environments through controlled optical attenuation, avoiding artifacts typically introduced by synthetic degradation, as shown in Fig.~\ref{fig5_dataset} (a).  

Fig.~\ref{fig5_dataset} (b) and (c) show the statistical characteristics and visualization results of the dataset, respectively. To characterize the illumination level and enhancement difficulty, we randomly crop spatially corresponding regions from each low-light RAW image and its paired normally exposed reference, and compute their mean intensity values. Analysis reveals that reference images are 300 $ \times $ to 500 
$ \times $ brighter than the RAW inputs on average, highlighting the severe photon-starvation regime and the extremely large gain and noise-suppression demand imposed on enhancement models. Illumination statistics further indicate that approximately 70\% of the scenes fall below 0.3~lux, confirming the extreme darkness covered by our dataset. Data are collected using a vehicle-mounted platform across diverse motion conditions. To balance motion sharpness with photon collection, we adopt a speed-aware exposure strategy: 1~ms for high-speed motion, 2~ms for moderate motion, and up to 5~ms in slow-motion scenes.

In summary, the dataset reflects real-world extreme low-illumination conditions with controlled camera motion and exposure scheduling to avoid blur. The dataset contains severely photon-limited inputs and large exposure gaps to the reference. Beyond low-level enhancement, we provide annotations for downstream high-level tasks, including object detection and semantic segmentation. This creates a realistic and challenging benchmark for low-light enhancement and event-based imaging. Dataset details are provided in the supplementary material.

\section{Experiments}
\label{sec:experiments}
\subsection{Datasets and Experimental Settings}
\noindent
\textbf{Datasets.}
We evaluate all methods on both synthetic and real datasets. For the synthetic evaluation, we use the RAW video dataset LLRVD \cite{fu2022lowv} and generate noisy events using V2E \cite{hu2021v2e}, named LLRVD-simu. For the real evaluation, since there is no publicly available event-RAW low-light dataset, we conduct experiments on the proposed REAL dataset.

\noindent
\textbf{Comparison Methods.}
We conduct a comprehensive comparison against four categories of SOTA methods, including sRGB-based, RAW-based, event-based, and event–frame hybrid approaches. The sRGB-based methods include LightenDiffusion~\cite{jiang2024lightendiffusion}, RetinexFormer~\cite{cai2023retinexformer}, SCI~\cite{ma2022toward}, LFPVs~\cite{xu2025learnable}, and SDSD~\cite{wang2021seeing}. The RAW-based methods consist of NoiseModelling~\cite{li2025noise}, LED~\cite{jin2023lighting}, PAP + HB~\cite{zhang2021rethinking}, and BRVE~\cite{zhang2024binarized}. The event-based methods include E2VID+~\cite{stoffregen2020reducing} and NER-Net~\cite{liu2024seeing}. The hybrid methods comprise EvLight~\cite{liang2024towards}, EvLowLight~\cite{liang2023coherent}, and ELEDNet~\cite{kim2024towards}. For methods that take sRGB images as input, RAW images are converted into sRGB space using the camera's recorded ISP parameters. To ensure fairness, all methods are fine-tuned separately on both the LLRVD-simu and REAL datasets.

\begin{table}
	\centering
	\caption{Quantitative comparison on LLRVD and REAL datasets.}
	\resizebox{1.0\linewidth}{!}{
		\begin{tabular}{llccccccc}
			\toprule
			\multirow{2.4}{*}{\textbf{Input}} & \multirow{2.4}{*}{\textbf{Method}} & \multicolumn{3}{c}{\textbf{LLRVD-simu}} & \multicolumn{3}{c}{\textbf{REAL}} \\
			\cmidrule(r){3-5} \cmidrule(l){6-8}
			& & PSNR↑ & SSIM↑ & LPIPS↓ & PSNR↑ & SSIM↑ & LPIPS↓ \\
			\midrule
			\multirow{5}{*}{\centering sRGB Only}
			& SDSD~\cite{wang2021seeing}             & 16.52 & 0.350 & 0.429 & 17.11 & 0.295 & 0.520 \\
			& SCI~\cite{ma2022toward}               & 17.71 & 0.444 & 0.586 & 21.26 & 0.576 & 0.340 \\
			& RetinexFormer~\cite{cai2023retinexformer} & 20.92 & 0.768 & 0.325 & 21.38 & 0.456 & 0.411 \\
			& LightenDiffusion~\cite{jiang2024lightendiffusion} & 21.64 & 0.818 & 0.265 & 22.19 & 0.714 & 0.282 \\
			& LFPVs~\cite{xu2025learnable}           & 17.86 & 0.626 & 0.406 & 20.09 & 0.637 & 0.312 \\
			\midrule
			\multirow{4}{*}{\centering RAW Only}
			& PAP + HB~\cite{zhang2021rethinking}     & 27.15 & 0.821 & 0.137 & 22.47 & 0.723 & 0.309 \\
			& LED~\cite{jin2023lighting}             & 25.78 & 0.823 & 0.153 & 16.48 & 0.628 & 0.364 \\
			& BRVE~\cite{zhang2024binarized}         & 27.58 & 0.817 & 0.137 & 21.87 & 0.717 & 0.334 \\
			& RID~\cite{li2025noise}                 & 26.76 & 0.825 & 0.127 & 22.72 & 0.729 & 0.258 \\
			\midrule
			\multirow{2}{*}{\centering Event Only}
			& E2VID+~\cite{stoffregen2020reducing}   & 12.53 & 0.512 & 0.342 & 14.78 & 0.472 & 0.368 \\
			& NER-Net~\cite{liu2024seeing}           & 14.78 & 0.655 & 0.236 & 15.93 & 0.601 & 0.320 \\
			\midrule
			\multirow{3}{*}{\centering Event-sRGB}
			& EvLowLight~\cite{liang2023coherent}     & 18.85 & 0.756 & 0.303 & 20.20 & 0.674 & 0.323 \\
			& EvLight~\cite{liang2024towards}         & 17.06 & 0.677 & 0.291 & 21.20 & 0.626 & 0.277 \\
			& ELEDNet~\cite{kim2024towards}           & 18.65 & 0.703 & 0.327 & 21.58 & 0.662 & 0.400 \\
			\midrule
			\multirow{1}{*}{\centering Event-RAW}
			& \textbf{NEC-Diff (Ours)} & \textbf{27.74} & \textbf{0.828} & \textbf{0.125} & \textbf{24.51} & \textbf{0.742} & \textbf{0.201} \\
			\bottomrule
		\end{tabular}
	}
	\label{tab1_quantative}
\end{table}

\subsection{Comparison Experiments}
\noindent
\textbf{Comparison on Synthetic Dataset.}
Fig.~\ref{fig6_qualitive} (top two rows) and Tab.~\ref{tab1_quantative} present the qualitative and quantitative comparisons on the LLRVD-simu dataset. Given the relatively simple noise distribution in simulated data, RAW-based methods yield superior performance on this dataset. In contrast, sRGB- and event-only approaches suffer from artifacts in extremely dark or smooth regions. Compared to other SOTA methods, NEC-Diff delivers balanced luminance and sharp edges for enhanced image quality.

\noindent
\textbf{Comparison on Real Dataset.}
Fig.~\ref{fig6_qualitive} (bottom two rows) and Tab.~\ref{tab1_quantative} show the comparison results on the REAL dataset. This dataset closely reflects authentic, intricate low-light scenarios. sRGB-based methods exhibit pronounced noise residuals, whereas RAW-based approaches effectively mitigate image noise but incur luminance imbalances and edge blurring. Event-only and hybrid methods falter in reconstructing plausible colors. In stark contrast, NEC-Diff exhibits marked superiority across multiple dimensions of visual quality.

\subsection{Ablation Study and Discussion}
\begin{table}[t]
	\setlength{\abovecaptionskip}{5pt}
	\setlength{\belowcaptionskip}{-5pt}
	\footnotesize
	\centering
	\caption{Ablation study of each module on the REAL dataset. }
	\resizebox{0.42\textwidth}{!}{
		\begin{tabular}{ccc|ccc}
			\Xhline{1px}
			ECNS & SRIE & CAD & PSNR~$\uparrow$ & SSIM~$\uparrow$ & LPIPS~$\downarrow$ \\
			\hline
			& \checkmark & \checkmark & 21.06 & 0.653 & 0.278 \\
			\checkmark & & \checkmark & 23.24 & 0.698 & 0.243 \\
			\checkmark & \checkmark & & 22.53 & 0.671 & 0.265 \\
			\checkmark & \checkmark & \checkmark & \textbf{24.51} & \textbf{0.742} & \textbf{0.201} \\
			\Xhline{1px}
	\end{tabular}}
	\label{tab:ablation}
\end{table}

\begin{figure}
	\centering
	\includegraphics[width=1.0\linewidth]{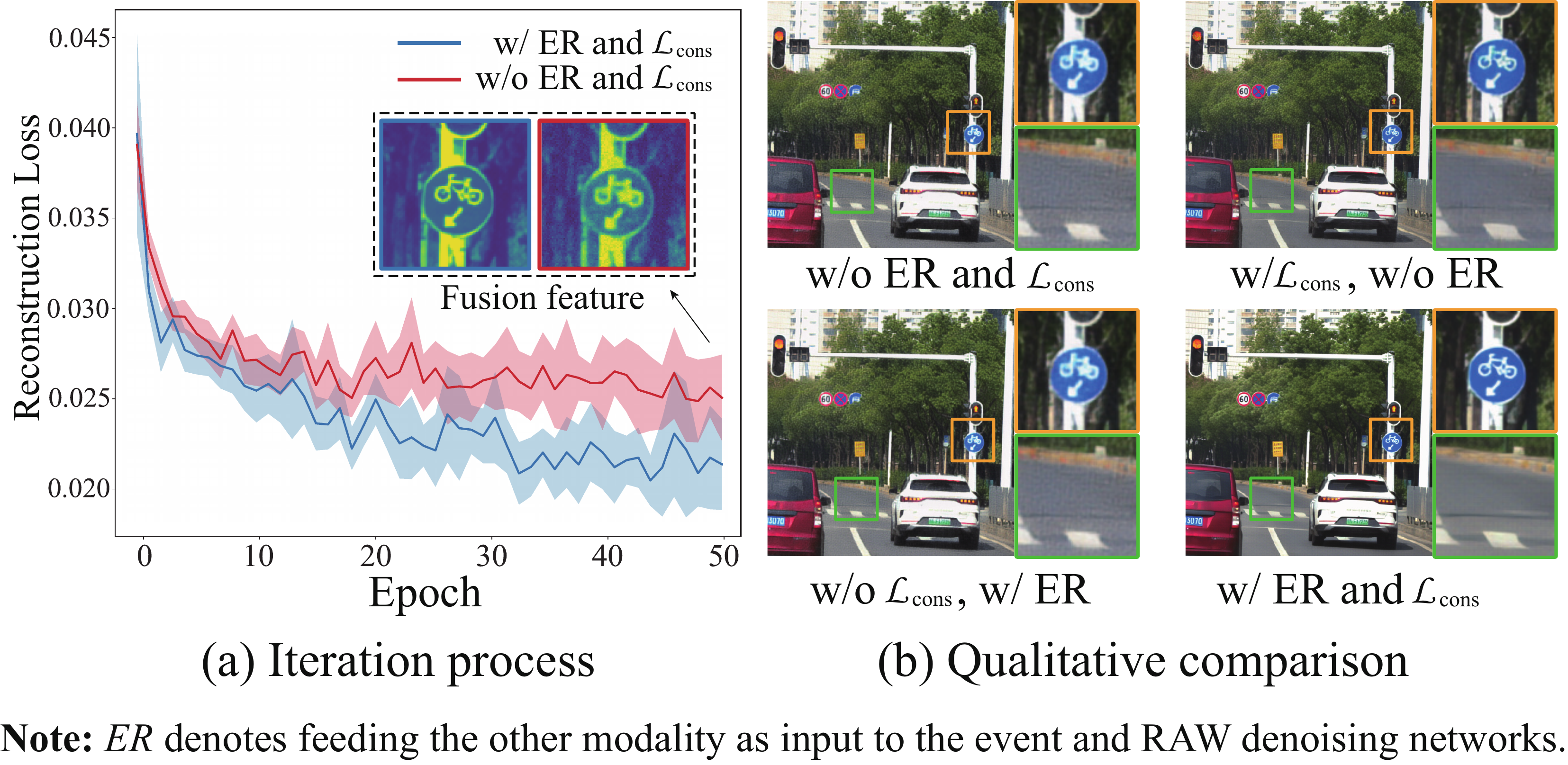}
	\caption{Effectiveness of ECNS. (a) ECNS effectively enhances the quality of reconstructed details. (b) Effects of cooperative denoising and consistency loss.}
	\label{fig7_ablation_ECNS}
\end{figure}

\noindent
\textbf{Effectiveness of Collaborative Noise Suppression.}
We evaluate the effectiveness of the ECNS module. As shown in Fig.~\ref{fig7_ablation_ECNS} (a), the red curve denotes denoising without cross-modal input or consistency loss, while the blue curve represents the full ECNS. The complete module consistently achieves lower reconstruction loss and yields finer textures during training.
As illustrated in Fig.~\ref{fig7_ablation_ECNS} (b), single-modal denoising leads to blurred textures, and using only cross-modal input or consistency loss brings limited gains. When both are jointly applied, ECNS effectively suppresses noise while preserving fine details, significantly improving image quality. Tab.~\ref{tab:ablation} further demonstrates the importance of the ECNS, as removing it leads to a 3.45 dB drop in PSNR.

\begin{figure}
	\centering
	\includegraphics[width=1.0\linewidth]{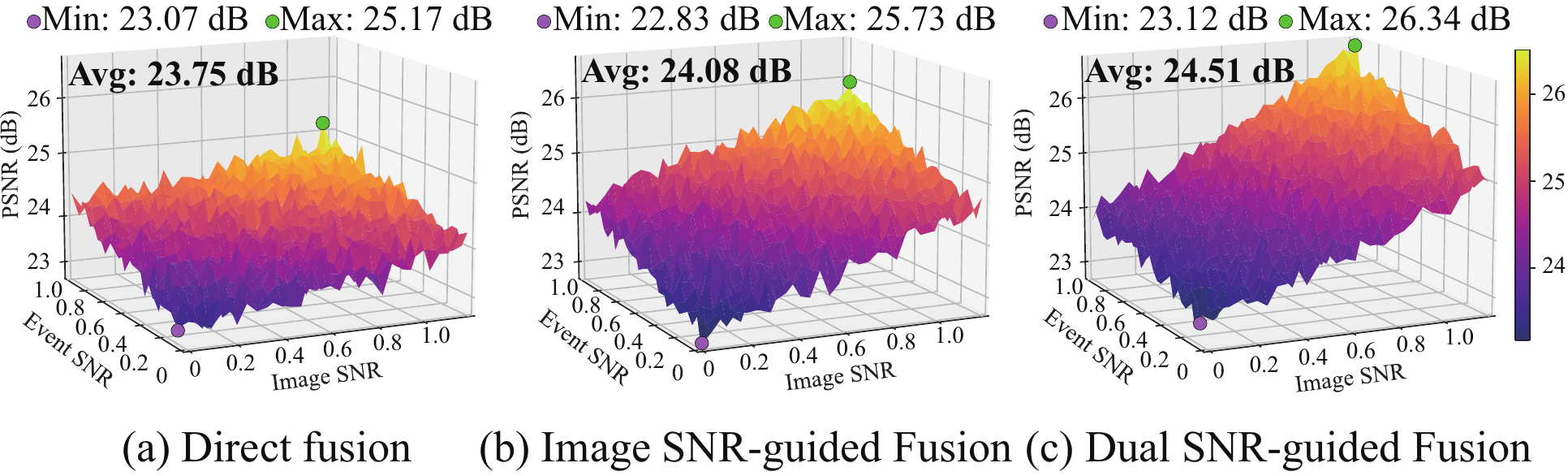}
	\caption{Effectiveness of SNR-guided fusion. Compared with (a) direct fusion and (b) image SNR-guided fusion, (c) dual SNR-guided fusion achieves the best performance.}
	\label{fig8_ablation_SNR}
\end{figure}

\noindent
\textbf{Effectiveness of SNR-guided fusion}
We analyze the effectiveness of different fusion strategies by statistically evaluating the event SNR, image SNR, and reconstruction quality on the REAL test dataset, as shown in Fig.~\ref{fig8_ablation_SNR}. Compared with direct fusion, the image SNR-guided fusion performs better when the image quality is relatively high, but degrades in regions where both the image and event SNRs are low. This is because the strategy tends to rely more on event features when the image quality deteriorates, thereby discarding the weak yet useful information that may still exist in the image. In contrast, the dual SNR-guided fusion strategy jointly considers the reliability of both modalities and achieves consistent improvements in all regions, increasing the average PSNR by 0.76 dB and 0.43 dB over the direct and image-guided strategies, respectively.

\begin{figure}
	\centering
	\includegraphics[width=1.0\linewidth]{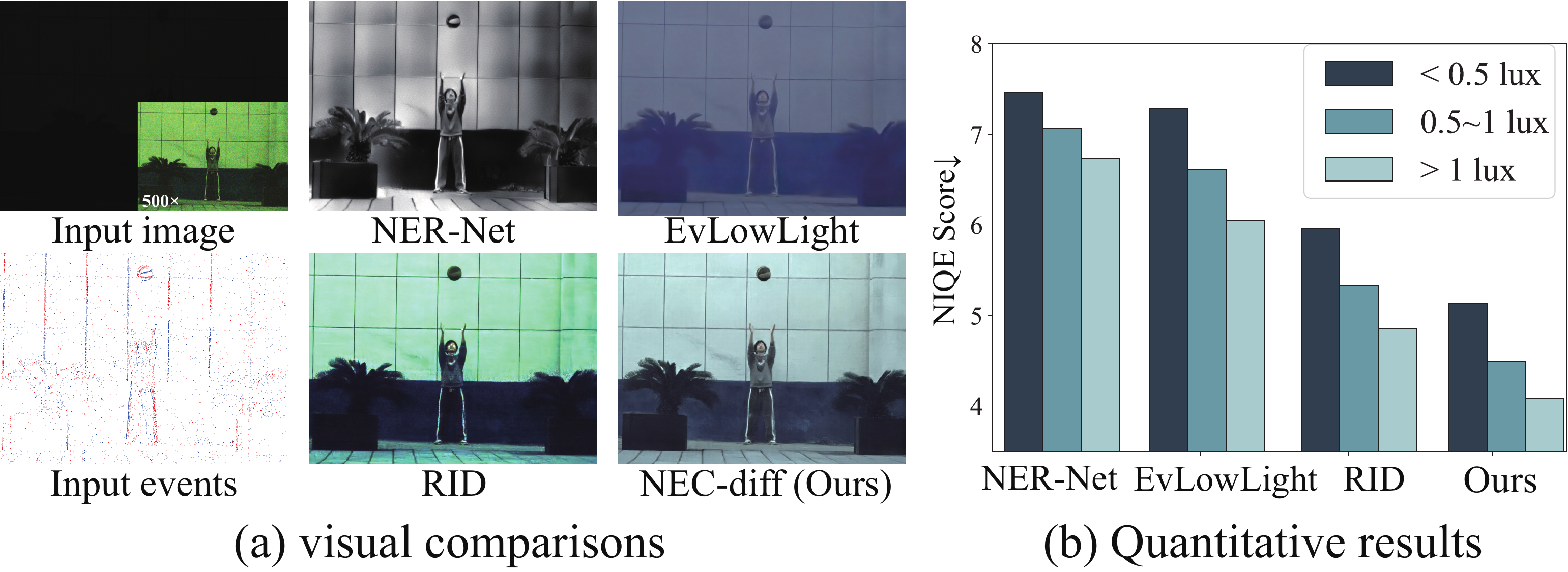}
	\caption{(a) Visual comparisons and (b) quantitative results in dynamic scenes under extremely low-light nighttime conditions.}
	\label{fig9_analysis_real}
\end{figure}

\noindent
\textbf{Generalization for Unseen Nighttime Scenes.}
We evaluated six real nighttime sequences with moving objects. Fig.~\ref{fig9_analysis_real} (a) shows our method yields superior texture and color fidelity, confirming strong generalization to unseen data. Quantitatively, Fig.~\ref{fig9_analysis_real} (b) reports NIQE scores \cite{mittal2012making} across varying illuminations, where NEC-Diff consistently outperforms all competitors.

\noindent
\textbf{Limitation and Future Work.}
While the intensity consistency loss effectively enhances denoising for both modalities, learning the event threshold $C$ from data implies that varying thresholds in practical applications may degrade noise suppression accuracy and limit generalization. To address this, future work will explore test-time adaptation \cite{cho2024tta} to fine-tune parameters during inference, enabling robust adaptation to threshold-induced distribution shifts.

\section{Conclusion}

In this work, we introduce NEC-Diff, a diffusion-based hybrid imaging framework that leverages the complementary strengths of event and RAW images for robust reconstruction in extreme darkness. It incorporates a physics-driven denoising constraint via illumination correlations between modalities, enabling reliable structure extraction from severely degraded signals. Additionally, SNR-guided adaptive fusion dynamically balances contributions, allowing the diffusion process to prioritize trustworthy data and preserve fine details. The proposed method significantly outperforms state-of-the-art approaches. We will release a large-scale paired dataset of low-light events, RAW images, and high-quality GTs to the community, facilitating broader research in extreme low-light imaging.

\clearpage
\newpage

\noindent
\textbf{Acknowledgments.}
This work was supported by the National Natural Science Foundation of China under Grant U24B20139, National Key Research and Development Program of China under Grant 2024YFB3909901, the Hubei Province Science Foundation of Distinguished Young Scholars under Grant JCZRJQ202500097.

{
    \small
    \bibliographystyle{ieeenat_fullname}
    \bibliography{main}
}


\end{document}